\pdfoutput=1

\documentclass[11pt]{article}

\usepackage{ACL2023}

\usepackage{times}
\usepackage{latexsym}

\usepackage[T1]{fontenc}

\usepackage[utf8]{inputenc}

\usepackage{microtype}

\usepackage{inconsolata}
\usepackage{amsmath,amsthm,amssymb,amsfonts}
\usepackage{graphicx}
\usepackage{multirow}
\usepackage{booktabs}

\newcommand{\methodname}{RepCL}

\title{RepCL: Exploring Effective Representation for Continual Text Classification}

\author{
Yifan Song$^1$$\footnotemark[1]$ \quad Peiyi Wang$^1$$\footnotemark[1]$ \quad Dawei Zhu$^1$ \quad Tianyu Liu$^2$ \\ \textbf{Zhifang Sui$^1$ \quad Sujian Li$^1$}
\\ 
$^1$ School of Computer Science, Peking University  \\
$^2$ Tencent Cloud Xiaowei \\
 {\{yfsong, szf, lisujian\}@pku.edu.cn}; {wangpeiyi9979@gmail.com}\\
 {\{rogertyliu\}@tencent.com;} 
}

\begin{document}
\maketitle

\renewcommand{\thefootnote}{\fnsymbol{footnote}}
\begin{abstract}



Continual learning (CL) aims to constantly learn new knowledge over time while avoiding catastrophic forgetting on old tasks.
In this work, we focus on continual text classification under the class-incremental setting.
Recent CL studies find that the representations learned in one task may not be effective for other tasks, namely representation bias problem.
For the first time we formally analyze representation bias from an information bottleneck perspective and suggest that exploiting representations with more class-relevant information could alleviate the bias.
To this end, we propose a novel replay-based continual text classification method, \methodname{}.
Our approach utilizes contrastive and generative representation learning objectives to capture more class-relevant features. 
In addition, \methodname{} introduces an adversarial replay strategy to alleviate the overfitting problem of replay.
Experiments demonstrate that \methodname{} effectively alleviates forgetting and achieves state-of-the-art performance on three text classification tasks.

\end{abstract}

\footnotetext[1]{Equal contribution.}
\renewcommand{\thefootnote}{\arabic{footnote}}

\section{Introduction}

\begin{figure}[t]
    \centering
    \includegraphics[width=\linewidth]{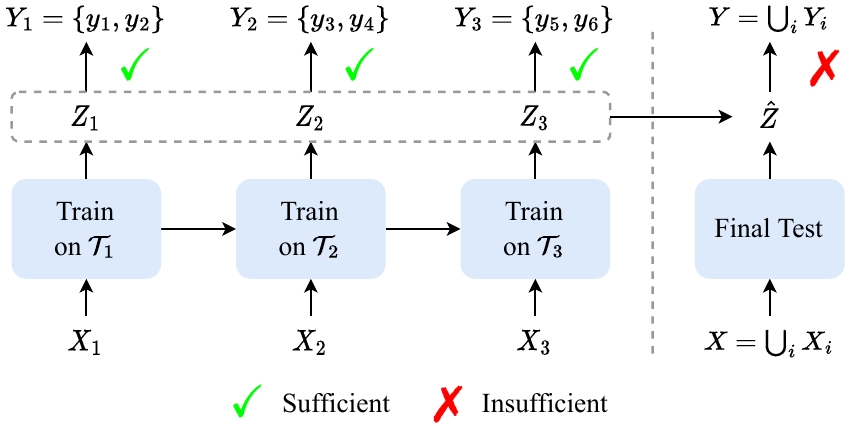}
    \caption{
    Illustration for continual text classification with three tasks where each task involves two new classes.
    $X_i$, $Y_i$, and $Z_i$ denote input sentences, new classes and learned representations for $i$-th task $\mathcal{T}_i$ respectively.
    Although the representations $Z_i$ learned in $\mathcal{T}_i$ are sufficient for classifying 
    $Y_i$, they are ineffective to distinguish all seen classes $Y$ in the final test.
    }
    \label{fig:intro}
\end{figure}

Continual learning (CL) enables conventional static natural language processing models to constantly gain new knowledge from a stream of incoming data \citep{lamol,cl_nlp_survey}.
In this paper, we focus on continual text classification, which is formulated as a class-incremental problem, requiring the model to learn from a sequence of class-incremental tasks \citep{huang-etal-2021-continual}.
Figure \ref{fig:intro} gives an illustrative example of continual text classification.
The model needs to learn to distinguish some new classes in each task and is eventually evaluated on all seen classes.
Like other CL systems, the major challenge of continual text classification is catastrophic forgetting \citep{cl_survey}: after new tasks are learned, performance on old tasks may degrade dramatically.

The earlier work in the CL community mainly attributes catastrophic forgetting to the corruption of the learned representations as new tasks arrive  and various methods have been introduced to retain or recover previously learned representations \citep{ewc,icarl,packnet,cl_survey}.
Recently, some studies find that, under the sequential task setting, CL models tend to only capture useful features for the current task and cannot learn effective representations with consideration of the entire CL process, which is named as representation bias and maybe the main reason for catastrophic forgetting \citep{co2l,ocm,aca}.
To alleviate representation bias, \citet{ocm} propose the  InfoMax model to greedily preserve the input features as much as possible and \citet{aca} exploit an adversarial class augmentation mechanism.
However, both InfoMax  and heuristic augmentation will introduce excessive noise into representations which may harm the generalization \citep{good_view_for_cl}.



To explore reasonable and effective representations, in this paper, for the first time we formally analyze the representation bias problem from the information bottleneck (IB) \citep{ib1,ib4} perspective.
Specifically, we formulate the learning process of CL as a trade-off between representation compression and preservation, and empirically measure the mutual information of the learned representations.
Then we derive the following conclusions that, (1) in each individual task, due to the compression effect of IB, CL models discard features irrelevant to current task and cannot learn sufficient representations for cross-task classes;
(2) effective representations for CL should ``globally'' balance feature preservation and task-irrelevant information compression for the entire continual task.
Since it is impossible to foresee future classes and identify crucial features for future tasks,  we suppose that the key problem is how to design representation learning objectives for capturing more class-relevant features.

Based on our analysis, we propose a replay-based continual text classification method \methodname{}.
\methodname{} utilizes both contrastive and generative representation learning objectives to learn 
class-relevant features from each individual task.
The contrastive objective explicitly maximizes the mutual information between representations of instances from the same class.
The generative objective instructs the representation learning process through reconstructing corrupted sentences from the same class, which implicitly makes the model capture more class-relevant features.
In addition, to better protect the learned representations, 
\methodname{} also incorporates an adversarial replay mechanism to alleviate the overfitting problem of the replay.

Our contributions are summarized as follows: 
(\textit{i}) We formally analyze representation bias in CL from an information bottleneck perspective and suggest that learning more class-relevant features will alleviate the bias.
(\textit{ii}) We propose a novel continual text classification method \methodname{}, which exploits contrastive and generative representation learning objectives to 
handle the representation bias. 
(\textit{iii}) Experimental results on three text classification tasks show that \methodname{} learns more effective representations and significantly outperforms state-of-the-art methods.

\section{Related Work}

\paragraph{Continual Learning}
Continual Learning (CL) studies the problem of continually learning knowledge from a sequence of tasks without the need to retrain from scratch \cite{cl_survey}.
The major challenge of CL is catastrophic forgetting.
Previous CL works mainly attribute catastrophic forgetting to the corruption of learned knowledge.
To this end, three major families of approaches have been developed.
\textit{Replay-based} methods \citep{icarl,gdumb} save a few previous task instances in a memory module and retrain on them while training of new tasks.
\textit{Regularization-based} methods \citep{ewc, mas} introduce an extra regularization term in the loss function to consolidate previous knowledge.
\textit{Parameter-isolation} methods \citep{packnet} dynamically expand the network and dedicate different model parameters to each task.
Despite the simplicity, replay-based methods have been proven to be effective \citep{der,wang-etal-2019-sentence}.
However, replay-based methods suffer from overfitting the few stored data.
In this paper, we introduce an adversarial replay strategy to alleviate the overfitting problem of replay.


\paragraph{Representation Learning in CL} 

Most previous work in CL focuses on retaining or recovering learned representations, whereas a few recent studies 
find that CL models suffer from representation bias: CL models tend to only capture useful features for the current task and cannot learn effective representations with consideration of the entire CL process \citep{ocm,aca}.
To mitigate the bias, \citet{aca} design two adversarial class augmentation strategies for continual relation extraction.
\citet{ocm} introduce a mutual information maximization method to preserve the features as much as possible for online continual image classification.
However, both heuristic data augmentation and InfoMax method may introduce task-irrelevant noise, which could lead to worse generalization on the task \citep{good_view_for_cl}.
Unlike previous work, in this paper, we formally analyze representation bias from the information bottleneck perspective and propose a more reasonable training objective that makes a trade-off between representation compression and preservation ability.

\section{Task Formulation}
In this work, we focus on continual learning for a sequence of $k$ class-incremental text classification tasks $(\mathcal{T}_1, \mathcal{T}_2, ..., \mathcal{T}_k)$. 
Each task $\mathcal{T}_i$ has its dataset $\mathcal{D}_i=\{(x_i^{(j)},y_i^{(j)})\}$, where $(x_i^{(j)},y_i^{(j)})$ is an instance of current task and is sampled from an individually i.i.d. distribution $p(X_i,Y_i)$.
Different tasks $\mathcal{T}_i$ and $\mathcal{T}_j$ have disjoint label sets $Y_i$ and $Y_j$.
The goal of CL is to continually train the model on new tasks to learn new classes while avoiding forgetting  previously learned ones.
From another perspective, if we denote $X=\cup_i X_i$ and $Y=\cup_i Y_i$ as the input and output space of the entire CL process respectively, continual learning is aiming to approximate a holistic distribution $p(Y|X)$ from a non-i.i.d data stream.

The text classification model $F$ is usually composed of two modules: the encoder $f$ and the classifier $\sigma$.
For an input $x$, we get the corresponding representation $z=f(x)$, and use the logits
$\sigma\left(z\right)$ to compute loss and predict the label.

\section{Representation Bias in CL}

Previous work \citep{co2l,ocm,aca,xia2023enhancing} reveals that representation bias is an important reason for catastrophic forgetting.
In this section, we analyze the representation bias problem from an information bottleneck (IB) perspective and further discuss 
what representations are effective for CL.

\subsection{Information Bottleneck}


We first briefly introduce the background of information bottleneck in this section.
Information bottleneck formulates the goal of deep learning as an information-theoretic trade-off between representation compression and preservation \citep{ib4,ib3}.
Given the input $\mathcal{X}$ and the label set $\mathcal{Y}$, one model is built to  learn the representation $\mathcal{Z}=\mathcal{F}(\mathcal{X})$ of the encoder $\mathcal{F}$.
The learning procedure of the model is to minimize the following Lagrangian: 
\begin{equation}
\label{eq:ib}
I(\mathcal{X};\mathcal{Z})-\beta I(\mathcal{Z};\mathcal{Y}),
\end{equation}
where $I(\mathcal{X};\mathcal{Z})$ is the mutual information (MI) between $\mathcal{X}$ and $\mathcal{Z}$, and $\beta$ is a trade-off hyperparameter.
With information bottleneck, the model will learn \textit{minimal sufficient representation} $\mathcal{Z}^*$ \citep{ib2} of $\mathcal{X}$ corresponding to $\mathcal{Y}$:
\begin{align}
&\mathcal{Z}^*= \arg \min_{\mathcal{Z}} I(\mathcal{X}; \mathcal{Z}) \\
&\mathrm{s.t.}\ I(\mathcal{Z}; \mathcal{Y}) =  I(\mathcal{X}; \mathcal{Y}).
\end{align}
Minimal sufficient representation is important for supervised learning, because it retains as little about input as possible to simplify the role of the classifier and improve generalization, without losing information about labels.

\subsection{Representation Bias: the IB Perspective}
\label{sec:reason}

In this section, we investigate representation bias from the IB perspective.
Continual learning is formulated as a sequence of individual tasks $(\mathcal{T}_1, \mathcal{T}_2, ..., \mathcal{T}_k)$. 
For $i$-th task $\mathcal{T}_i$, the model aims to approximate distribution $p(Y_i|X_i)$.
According to IB, if the model $F=\sigma\circ f$ converges, the learned hidden representation $Z_i=f(X_i)$ will be minimal sufficient for $\mathcal{T}_i$:
\begin{align}
&Z_i = \arg \min_{Z_i} I\left(X_i;Z_i\right) \\
&\mathrm{s.t.}\ I\left(Z_i;Y_i\right)=I(X_i;Y_i),
\end{align}
which ensures the performance and generalization ability of the current task.
Nevertheless,
the minimization of the compression term will bring potential risks: features that are useless in the current task but crucial for other tasks will be discarded.

For the entire continual learning task with the holistic target distribution $p(Y|X)$, the necessary condition to perform well is that the representation $Z$ is sufficient for $Y$: $I(Z;Y)=I(X;Y)$.
However, as some crucial features are compressed, the combination of minimal sufficient representations for each task $Z=\cup_i Z_i$ may be insufficient:
\begin{equation}
I\left(Z;Y\right)<I(X;Y).
\end{equation}
Therefore, from the IB perspective, representation bias can be reformulated as: due to the compression effect of IB, the learned representations in each individual task may be insufficient for the entire continual task.


\begin{table}[t]
    \setlength\tabcolsep{3pt}
    \small
    \centering
    \scalebox{0.98}{
        \begin{tabular}{lcccc}
         \toprule
            \multirow{2}*{\textbf{Models}} & \multicolumn{2}{c}{\textbf{FewRel}} & \multicolumn{2}{c}{\textbf{MAVEN}} \\
            \cmidrule(r){2-3} \cmidrule(r){4-5}
            ~ & $I(X_1;Z_1)$ & $I(Z;Y)$ & $I(X_1;Z_1)$ & $I(Z;Y)$ \\
            \midrule
            Supervised & 2.42 & 2.45 & 3.50 & 2.42 \\
            \midrule
            CRL & 2.12 & 2.18 & 3.12 & 2.30 \\
            CRECL & 2.20 & 2.31 & 3.01 & 2.36 \\
            FEA & 2.35 & 2.34 & 3.17 & 2.37 \\
            \bottomrule
    \end{tabular}}
    \caption{
    Mutual information comparison between supervised learning and strong CL baselines on FewRel and MAVEN datasets.
    We use $I(X;Z)$ to measure how much features of input $X$ representation $Z$ preserves.
    To exclude the impact of representation corruption, we instead estimate $I(X_1;Z_1)$ after CL models finish $\mathcal{T}_1$.
    $I(Z;Y)$ measures whether the learned representation is sufficient for the entire continual task.
    }
    \label{tab:ib}
\end{table}


To confirm our analysis, we use supervised learning on all data as the baseline, and compare MI between supervised learning with several strong CL baselines.
Concretely, we use MINE \citep{mine} as the MI estimator and conduct experiments on FewRel and MAVEN dataset\footnote{See Section \ref{sec:exp_setup} for details of CL baselines and datasets.}.
First, we measure $I(X;Z)$ to estimate the amount of features preserved by the representation $Z$.
However, previously learned representations will be corrupted after the model learns new tasks, which will make our estimation inaccurate.
To exclude the impact of representation corruption, we instead estimate $I(X_1;Z_1)$ on $\mathcal{T}_1$'s test set.
Second, to measure whether learned representations are sufficient for the entire continual task, we compare $I(Z;Y)$ on the final test set with all classes.
As shown in Table \ref{tab:ib}, both $I(X_1;Z_1)$ and $I(Z;Y)$ of three CL models are significantly lower than supervised learning, indicating that the CL model tends to compress more information due to the individual task setting and the representations learned in CL are insufficient for the entire continual task.


\subsection{What are Effective CL Representations?}
\label{sec:obj}

Since minimization of the compression term $I(\mathcal{X};\mathcal{Z})$ in IB leads to representation bias in CL, the most straightforward solution is to defy it when learning new tasks.
Specifically, in $i$-th task $\mathcal{T}_i$, we can maximize $I\left(X_i;Z_i\right)$ and force the representation to retain information about input as much as possible \citep{ocm} (we will omit subscripts $i$ for brevity of notation).
However, \citet{good_view_for_cl} find that the InfoMax principle may introduce task-irrelevant noisy information, which could lead to worse generalization.

Ideally, the effective representation learning objective of CL should be a ``global'' IB trade-off between feature preservation and task-irrelevant information compression for the entire continual task.
Due to the sequential task setting of CL, we cannot foresee future tasks and identify crucial features for the entire continual task in advance.
Nevertheless, if we can capture the ``essence'' features closely relevant to a specific class, then the representation is discriminative with any other classes.  
On the other hand, without knowing future tasks, any information shared by instances of the same class are potentially useful for the entire CL process.
Therefore, we propose that a more effective representation learning objective for new tasks is to learn more class-relevant features,
which can improve the sufficiency of the representations with as little impact on the generalization of the current task as possible.
The empirical results in Section \ref{sec:repr_analyze} also show that learning more class-relevant features has better performance than directly maximizing $I(X;Z)$.

\section{Methodology}
\label{sec:method}

\begin{figure*}[t]
    \centering
    \scalebox{0.99}{
        \includegraphics[width=\linewidth]{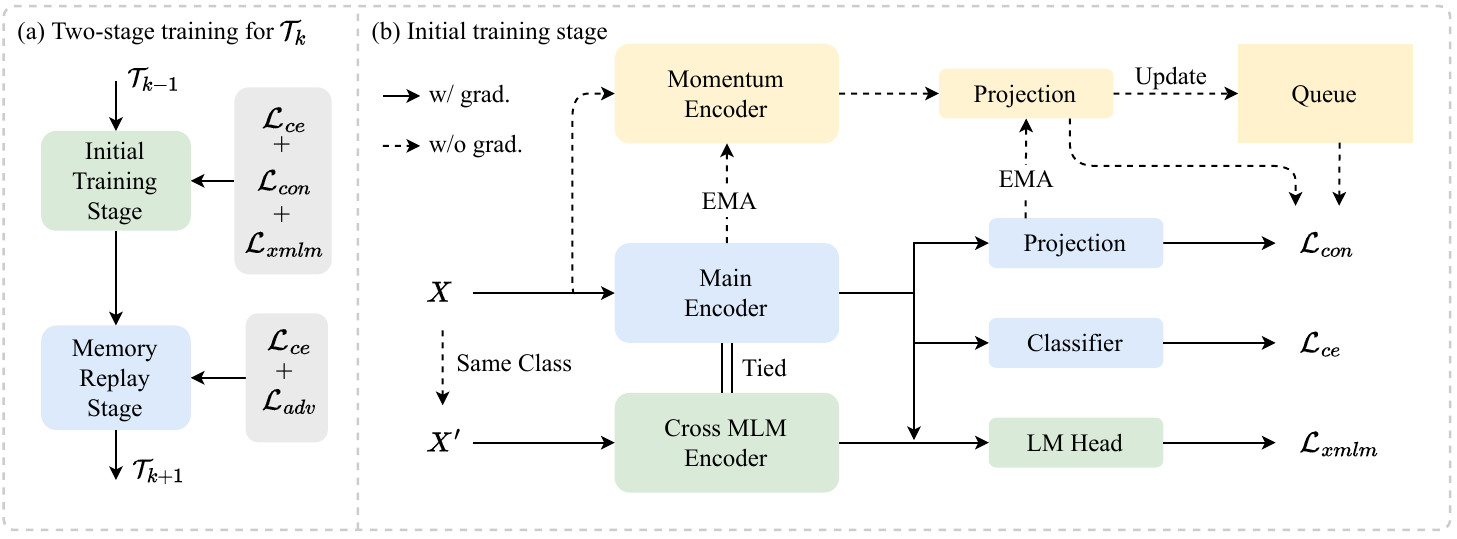}
    }    
    \caption{
    (a) A demonstration for \methodname{}, which is based on the two-stage training backbone.
    (b) Framework of initial training stage of InfoCL, which can be decomposed into (yellow) momentum branch, (blue) main branch, (green) XMLM branch.
    In the memory replay stage and at inference time, we discard momentum branch and XMLM branch, only preserving the main branch.
    }
    \label{fig:method}
\end{figure*}

Based on our analysis, we propose a novel replay-based continual learning method, \methodname{}, which helps the model learn more class-relevant features to better alleviate representation bias.

Following recent work \citep{ocl_survey,aca}, we adapt our method to the two-stage training backbone, which can effectively alleviate the classifier bias problem for replay-based methods.
Concretely, the training process on a task is divided into the initial training and memory replay stages.
In the initial training stage, the model is trained with only current task data to learn the new task.
In the memory replay stage, we first select a few typical instances from new task data to update memory and then train the model with the balanced memory bank to recover previously learned knowledge.

The overall structure of \methodname{} is illustrated in Figure \ref{fig:method}.
In Section \ref{sec:supinfonce} and \ref{sec:xmlm}, we introduce contrastive and generative representation learning for the initial training stage to learn class-relevant representations and mitigate representation bias.
In Section \ref{sec:adv}, we propose an adversarial replay mechanism in the memory replay stage to better recover corrupted old representations.

\subsection{Contrastive Representation Learning}
\label{sec:supinfonce}

If we denote $Z$ and $Z^+$ as the representations corresponding to instances $X$ and $X^+$ with the same label, $I(Z;Z^+)$ measures the amount of class-relevant features contained in representations.
Due to the intractability of computing MI, 
we instead use contrastive loss SupInfoNCE \citep{supinfonce} as a proxy to maximize a lower bound on $I\left( f(X);f(X^+) \right)$ to learn more class-relevant features\footnote{For more details about SupInfoNCE, the connection with InfoNCE \cite{cpc}, and the comparison against supervised contrastive learning (SupCon) \citep{supcon}, see Appendix \ref{app:supinfonce}.}:
\begin{equation}
\label{eq:supinfonce2}
\resizebox{.95\hsize}{!}{$
\mathbb{E}_{p(x)}\left[ \mathbb{E}_{p(x^+|x)}\left[ g(x,x^+) - \mathbb{E}_{q(\mathcal{B})}\left[ \log\sum\limits_{x^-\in\mathcal{B}}\exp g(x,x^-) \right] \right] \right],
$}
\end{equation}
where $g(x_1,x_2)=f(x_1)\cdot f(x_2)/\tau$ computes the representation similarity of two instances, $x^+$ is the positive instance with the same label as $x$, $x^-$ is an instance of the current batch $\mathcal{B}$, and $\tau$ is the temperature hyperparameter.



As the number of negative instances increases, SupInfoNCE will be a tighter bound of MI \citep{cpc,poole}.
Therefore, to effectively enlarge the number of negative instances, following MoCoSE \citep{cao-etal-2022-exploring}, we build a contrastive learning model consisting of a two-branch structure and a queue.
The upper two branches of Figure \ref{fig:method}(b) depicts the architecture.
As illustrated, we add a projection layer on both of the main branch and the momentum branch.
For the queue, same as \citet{moco, cao-etal-2022-exploring}, it is updated with the output of momentum branch by first-in-first-out strategy.

As mentioned before, we use SupInfoNCE as contrastive representation learning objective:
\begin{equation}
\label{eq:supinfonce}
\resizebox{.95\hsize}{!}{$
\begin{aligned}
&\mathcal{L}_{\mathrm{con}}= \\
&-\sum_i \log \frac{\exp (z\cdot z^+_i /\tau)}{\exp (z\cdot z^+_i /\tau) + \sum_{z^-} \exp (z\cdot z^- /\tau)},
\end{aligned}
$}
\end{equation}
where $z$ is the anchor vector obtained by the main branch, $z_i^+$ refers to the vector obtained by the momentum branch or stored in the queue which has the same class label as the anchor, and $z^-$ are the negative instances in the momentum branch or in the queue which belong to different classes with anchor, $\tau$ is the temperature hyperparameter.

The model calculates $\mathcal{L}_{\mathrm{con}}$ and backwards to update the main branch.
The momentum branch truncates the gradient and is updated with exponential moving average (EMA) method during training.
Formally, denoting the parameters of the main and momentum branches as $\theta$ and $\theta'$, $\theta'$ is updated by:
\begin{equation}
\theta' \leftarrow \eta \theta' + (1-\eta)\theta,
\end{equation}
where $\eta$ is the EMA decay rate.


\subsection{Generative Representation Learning}
\label{sec:xmlm}

In Eq. \ref{eq:supinfonce}, to make SupInfoNCE an unbiased MI estimator, the negative instances $z^-$ should be sampled from the whole input space. 
However, in CL scenario, $z^-$ are still sampled from the current task, thus the effectiveness of contrastive learning may be limited. 
Inspired by recent generative sentence representation work \citep{yang-etal-2021-universal,paser}, we additionally introduce a novel cross masked language modeling (XMLM) training objective to encourage the learned representations to contain more class-relevant information.

The lower two branches in Figure \ref{fig:method}(b) illustrate the architecture of XMLM.
For instance $x$, we randomly sample another instance $x'$ with the same class label as $x$ from the current task's training set.
Then $x'$ is masked in a certain proportion, and fed into the XMLM encoder, which shares the same weights with the main branch encoder.
The masked language modeling task of $x'$ is aided by representation $f(x)$ output by the main encoder.
Specifically, we concatenate $f(x)$ with the hidden states of token \texttt{[MASK]} in $x'$ to predict the corresponding masked token and compute MLM loss $\mathcal{L}_{\mathrm{xmlm}}$.
Intuitively, if the representation $f(x)$ contains more class-relevant information, it will be helpful for the XMLM branch to recover the corrupted sentence from the same class.
We empirically compare XMLM with vanilla MLM in Section \ref{sec:ablation}.

When a new task comes, we first initialize a new momentum branch and XMLM branch.
Then in the initial training stage we optimize the model with the combination of cross-entropy loss, contrastive loss, and XMLM loss as objective:
\begin{equation}
\mathcal{L}_{\mathrm{init}} = \mathcal{L}_{\mathrm{ce}} + \lambda_1\mathcal{L}_{\mathrm{con}} + \lambda_2\mathcal{L}_{\mathrm{xmlm}},
\end{equation}
where $\lambda_1, \lambda_2$ are weighting coefficients.
In the memory replay stage and at inference time, we discard the momentum branch and XMLM branch, only using the output of the main branch classifier to predict the label of an instance.

\subsection{Adversarial Replay}
\label{sec:adv}
After the initial training stage, we first select and store typical instances for each class for replay.
Following \citet{cui-etal-2021-refining} and \citet{zhao-etal-2022-consistent}, for each class, we use K-means to cluster the corresponding representations, and the instances closest to the centroids are stored in the memory bank.

Then we use the instances of all seen classes in the memory bank to conduct the memory replay stage.
The replay strategy is widely used to recover corrupted representations, whereas its performance is always hindered by the overfitting problem due to the limited memory budget.
To alleviate overfitting and enhance the effect of recovering, we incorporate FreeLB \citep{freelb} adversarial loss into the supervised training objective:
\begin{equation}
\label{eq:freelb}
\resizebox{.9\hsize}{!}{$
\begin{aligned}
&\mathcal{L}_{\mathrm{adv}}= \\
&\min\limits_{\theta} \mathbb{E}_{(x,y)\sim \mathcal{M}} \left[ \frac{1}{K}\sum_{t=0}^{K-1} \max\limits_{\Vert\delta_t\Vert \le \epsilon} \mathcal{L}\left( F(x+\delta_t),y \right) \right].
\end{aligned}
$}
\end{equation}
Please refer to Appendix \ref{app:adv} for details about Eq. \ref{eq:freelb}.
Intuitively, FreeLB performs multiple adversarial attack iterations to craft adversarial examples,
which is equivalent to replacing the original batch with a $K$-times larger adversarial augmented batch.

The optimization objective in the memory replay stage is the combination of $\mathcal{L}_{\mathrm{adv}}$ and conventional cross entropy loss $\mathcal{L}_{\mathrm{ce}}$:
\begin{equation}
\mathcal{L}_{\mathrm{replay}} = \mathcal{L}_{\mathrm{ce}} + \mathcal{L}_{\mathrm{adv}}.
\end{equation}

\begin{table*}[t]
    \centering
    \scalebox{0.95}{
    \begin{tabular}{l|cc|cc|cc|cc}
     \toprule
     \textbf{Datasets} & \multicolumn{2}{c|}{\textbf{FewRel}} & \multicolumn{2}{c|}{\textbf{TACRED}} & \multicolumn{2}{c|}{\textbf{MAVEN}} & \multicolumn{2}{c}{\textbf{HWU64}} \\
     \midrule
      \textbf{Models} & \textbf{Acc $\uparrow$} & \textbf{FR $\downarrow$} & \textbf{Acc $\uparrow$} & \textbf{FR $\downarrow$} & \textbf{Acc $\uparrow$} & \textbf{FR $\downarrow$} & \textbf{Acc $\uparrow$} & \textbf{FR $\downarrow$}  \\
    \midrule
    IDBR \citep{huang-etal-2021-continual} & 68.9 &  30.4 & 60.1 & 35.3 & 57.3 & 34.2 & 76.1 & 19.0 \\
    KCN \citep{cao-etal-2020-incremental} & 76.0 & 23.2 & 70.6 & 22.3 & 64.4 & 29.0 & 81.9 & 13.5 \\
    KDRK \citep{yu-etal-2021-lifelong} & 78.0 & 18.4 & 70.8 & 22.8 & 65.4 & 28.3 & 81.4 & 14.0 \\
    EMAR \citep{han-etal-2020-continual} & 83.6 & 12.1 & 76.1 & 20.0 & 73.2 & 14.6 & 83.1 & 9.3 \\
    RP-CRE \citep{cui-etal-2021-refining} & 82.8 & 10.3 & 75.3 & 17.5 & 74.8 & 11.4 & 82.7 & 10.9 \\
    CRL \citep{zhao-etal-2022-consistent} & 83.1 & 11.5 & 78.0 & 18.0 & 73.7 & 11.2 & 81.5 & 9.9 \\
  CRECL \citep{hu-etal-2022-improving} & 82.7 & 11.6 & 78.5 & 16.3 & 73.5 & 13.8 & 81.1 & 9.8 \\
     FEA \citep{wang2022less} & 84.3 & 8.9 & 77.7 & 13.3 & 75.0 & 12.8 & 83.3 & 8.8 \\
    ACA \citep{aca} & 84.7 & 11.0 & 78.1 & 13.8 & -- & -- & -- & -- \\
    \midrule
    \methodname{} (Ours)& \bf{85.6} & \bf{8.7} & \bf{78.6}  & \bf{12.0} & \textbf{75.9} & \bf{10.7}  &\bf{84.8} & \bf{8.0} \\
     \bottomrule
    \end{tabular}
    }
    \caption{
    Accuracy (Acc) and forgetting rate (FR) on all seen classes after learning the final task. 
    We report the average result of $5$ different runs. The best results are in \textbf{boldface}. 
    ACA  is specially designed for continual relation extraction and cannot be adapted to other tasks. 
    CRECL  is a concurrent work to ours. 
    }
    \label{tab:main}
\end{table*}

\section{Experiments}

\subsection{Experiment Setups}
\label{sec:exp_setup}
\paragraph{Datasets}
To fully measure the ability of \methodname{}, we conduct experiments on 4 datasets for 3 different text classification tasks, including relation extraction, event classification, and intent detection.
For relation extraction, following previous work \citep{han-etal-2020-continual,cui-etal-2021-refining,zhao-etal-2022-consistent}, we use \textbf{FewRel} \citep{han-etal-2018-fewrel} and \textbf{TACRED} \citep{zhang-etal-2017-position}.
For event classification, following \citet{yu-etal-2021-lifelong} and \citet{plm_cl}, we use \textbf{MAVEN} \citep{wang-etal-2020-maven} to build our benchmark.
For intent detection, following \citet{msr}, we choose \textbf{HWU64} \citep{hwu64} dataset.
For the task sequence, we simulate 10 tasks by randomly dividing all classes of the dataset into 10 disjoint sets,
and the number of new classes in each task for FewRel, TACRED, MAVEN and HWU64 are 8, 4, 12, 5 respectively.
For a fair comparison, the result of baselines are reproduced on the same task sequences as our method.
Please refer to Appendix \ref{app:dataset} for details of these four datasets.

\paragraph{Evaluation Metrics}
Following previous work \citep{hu-etal-2022-improving, aca}, we use the average accuracy ({Acc}) on all seen tasks as our main metric.
In addition to, we also use the average forgetting rate ({FR}) \citep{rwalk} to quantify the accuracy drop of old tasks.
CL methods with lower forgetting rates have less forgetting on previous tasks.
The detailed computation of FR is given in Appendix \ref{app:forgetting_rate}.

\paragraph{Baselines}

We compare \methodname{} against the following baselines: IDBR \citep{huang-etal-2021-continual}, KCN \citep{cao-etal-2020-incremental}, KDRK \citep{yu-etal-2021-lifelong}, EMAR \citep{han-etal-2020-continual}, RP-CRE \citep{cui-etal-2021-refining}, CRL \citep{zhao-etal-2022-consistent}, CRECL \citep{hu-etal-2022-improving}, FEA \citep{wang2022less} and ACA \citep{aca}. 
Note that FEA \citep{wang2022less} can be seen as the two-stage training backbone of \methodname{}.
See Appendix \ref{app:baselines} for details of our baselines.

Some baselines are originally proposed to tackle one specific task.
For example, RP-CRE is designed for continual relation extraction.
We adapt these baselines to other tasks and report the corresponding results.
Although the continual image classification work \citet{co2l} and \citet{ocm} are also conceptually related to our work, their methods rely on specific image augmentation mechanisms and cannot be adapted to the continual text classification task. 

\paragraph{Implementation Details}
For \methodname{}, we use BERT$_{\mathrm{base}}$ \citep{devlin-etal-2019-bert} as the encoder following previous work \cite{cui-etal-2021-refining,aca}.
The learning rate of \methodname{} is set to 1e-5 for the BERT encoder and 1e-3 for other modules.
Hyperparameters are tuned on the first three tasks.
The memory budget for each class is fixed at 10 for all methods.
For all experiments, we use NVIDIA A40 and RTX 3090 GPUs and report the average result of 5 different task sequences. 
More implementation details can be found in Appendix \ref{app:details}.

\subsection{Main Results}

Table \ref{tab:main} shows the performance of \methodname{} and baselines on four datasets for three text classification tasks.
Due to space constraints, we only illustrate Acc after learning the final task and FR.
The complete accuracy of all 10 tasks can be found in Appendix \ref{app:main_exp}.
As shown, our proposed \methodname{} consistently outperforms all baselines (except CRECL in TACRED) with significance test $p<0.05$ and achieves new state-of-the-art results on all four benchmarks.
Furthermore, compared with baselines on all tasks, \methodname{} has a substantially lower forgetting rate, which indicates that our method can better alleviate catastrophic forgetting.
These experimental results demonstrate the effectiveness and universality of our proposed method.

\section{Analysis}

\subsection{Ablation Study}
\label{sec:ablation}

\begin{table}[t]
\centering
\small
\scalebox{1}{
    \begin{tabular}{lcccc}
    \toprule
    \textbf{Models} & \textbf{Few.} & \textbf{TAC.} & \textbf{MAV.} & \textbf{HWU.} \\
    \midrule
    \methodname{} & 85.6 & 78.6 & 75.9 & 84.8  \\
    \midrule
    w/o $\mathcal{L}_{\mathrm{con}}$ & 85.3 & 78.1 & 75.6 & 84.5  \\
    w/o $\mathcal{L}_{\mathrm{xmlm}}$ & 84.8 & 77.9 & 75.5  & 84.1 \\
    w/o $\mathcal{L}_{\mathrm{adv}}$ & 85.3 & 78.0 & 75.9 & 83.8  \\
    \midrule
    w/ MLM & 85.0 & 78.2 & 75.3 & 84.3 \\
    \bottomrule
    \end{tabular}
}
\caption{
Ablation study of \methodname{}.
``w/ MLM'' denotes replacing cross MLM objective with a conventional masked language modeling objective.
}
\label{tab:ablation}
\end{table}

We conduct an ablation study to investigate the effectiveness of different components of \methodname{}.
The results are shown in Table \ref{tab:ablation}.
We find that the three core mechanisms of \methodname{}, namely contrastive and generative objectives for learning new tasks, and adversarial replay for the replay stage, are conducive to the model performance.
Note the generative objective $\mathcal{L}_{\mathrm{xmlm}}$ is more effective than the contrastive objective $\mathcal{L}_{\mathrm{con}}$.
We attribute this to the fact that negative instances for contrastive learning are not sampled from the whole input space, which has been discussed in Section \ref{sec:xmlm}.

To better understand our proposed generative objective, we replace XMLM with the conventional MLM objective and conduct experiments.
Specifically, in the XMLM branch, we remove the concatenation of the main branch representation.
As shown, training with MLM leads to performance degradation, indicating that vanilla MLM cannot effectively guide the representations to contain more class-relevant information.

\subsection{Effective Representation Learning}
\label{sec:repr_analyze}

In Section \ref{sec:obj}, we suggest that learning more class-relevant features is a better training objective to alleviate representation bias in CL, which makes a trade-off between crucial feature preservation and noisy information compression.
Since FEA \citep{wang2022less} can be seen as the backbone of our method, we compare \methodname{} with FEA and its InfoMax variant FEA+InfoMax, which replaces our proposed contrastive and generative learning objectives with InfoNCE as \citet{gao-etal-2021-simcse} to maximize $I(X;Z)$ in the initial training stage.
To exclude the impact of representation corruption, we choose $\mathcal{T}_1$ and estimate $I(X_1;Z_1)$ and $I\left(Z_1;Z_1^+\right)$ of representations learned by different methods.

\begin{table}[t]
\centering
\small
\begin{tabular}{lccc}
\toprule 
\textbf{Models} & FEA & FEA+InfoMax & RepCL \\
\midrule
$I(X_1;Z_1)$ & 2.35 & 2.44 & 2.41 \\
$I(Z_1;Z_1^+)$  & 1.99 & 1.99 & 2.01\\
\midrule
Final Acc & 84.3 & 84.7 & 85.6 \\
\bottomrule
\end{tabular}
\caption{
$I(X_1;Z_1)$ and $I(Z_1;Z_1^+)$ of test set in $\mathcal{T}_1$ of FewRel.
Since the estimate process of $I(Z_1;Z_1^+)$ is not stable, we illustrate the estimator fitting curves in Figure \ref{fig:analyze}(a).
FEA+InfoMax introduces InfoNCE loss following \citet{gao-etal-2021-simcse} to maximise $I(X;Z)$.
We also report the final accuracy on all tasks for three methods.
}
\label{tab:analyze}
\end{table}

\begin{figure}[t]
    \centering
    \includegraphics[width=1\linewidth]{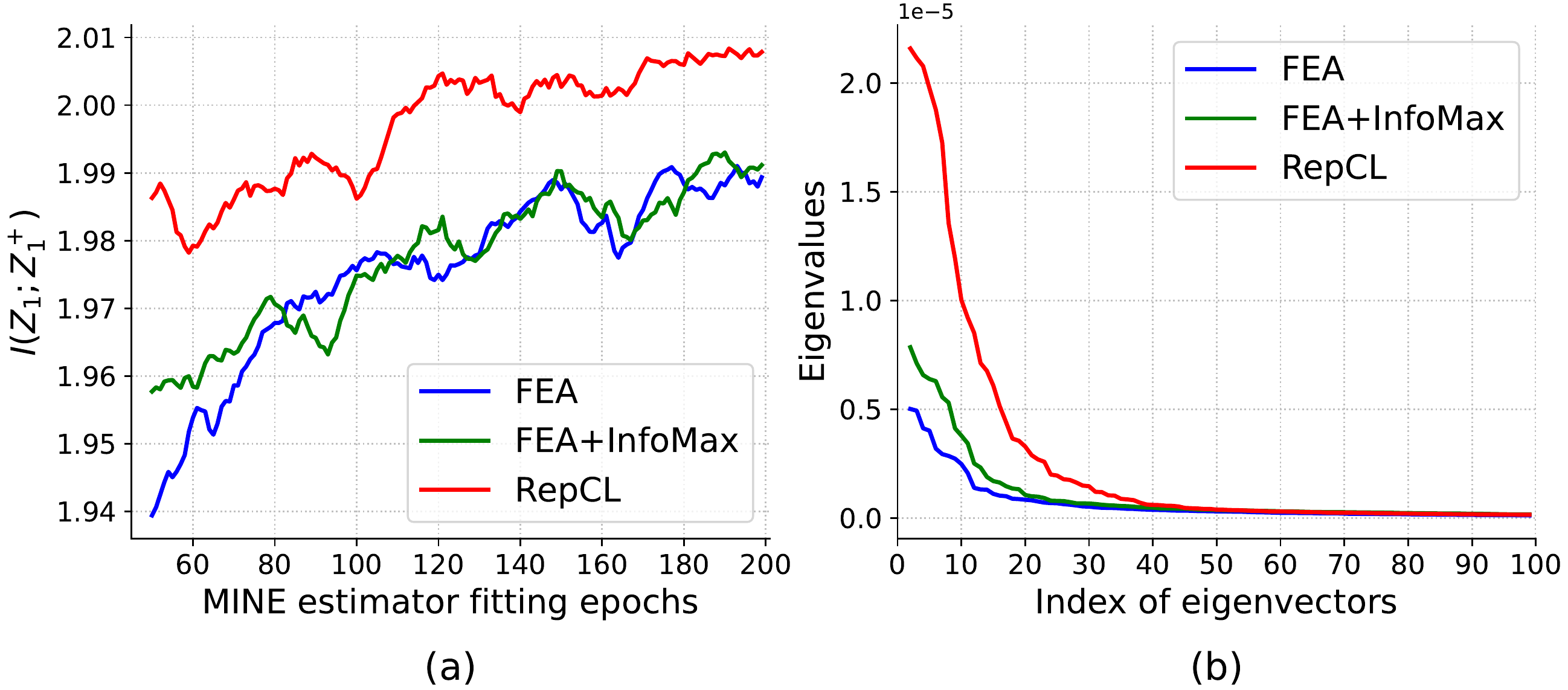}
    \caption{
    (a) MI estimator fitting curves of $I\left(Z_1;Z^+_1\right)$ of test set in $\mathcal{T}_1$ of FewRel.
    (b) Distributions of eigenvalue for the representations learned by baselines and \methodname{} on $\mathcal{T}_1$ of FewRel.
    Top-100 largest eigenvalues are shown.
    }
    \label{fig:analyze}
\end{figure}

The results are shown in Table \ref{tab:analyze} and Figure \ref{fig:analyze}(a).
Both FEA+InfoMax and \methodname{} have higher $I(X_1;Z_1)$ than FEA baseline, showing that they can learn more sufficient representations.
However, compared with \methodname{}, FEA+InfoMax has a lower $I(Z_1;Z_1^+)$ and worse final accuracy, indicating the InfoMax objective will introduce task-irrelevant information and lead to worse generalization.
In contrast, $I(Z_1;Z_1^+)$ and final accuracy of \methodname{} are higher than those of two baselines, demonstrating that our method can capture more class-relevant features and better alleviate representation bias.

Additionally, \citet{dual_aug} observed that representations with larger eigenvalues transfer better and suffer less forgetting.
Therefore, in Figure \ref{fig:analyze}(b), we also illustrate the distributions of eigenvalue for the representations learned in $\mathcal{T}_1$ of FewRel.
As shown, the eigenvalues of \methodname{} are higher than two baselines, which also demonstrates that our method can learn effective representations for CL.

\subsection{Influence of Memory Size}

Memory size is the number of memorized instances for each class, which is an important factor for the performance of replay-based CL methods.
Therefore, in this section, we study the impact of memory size on \methodname{}.
We compare the performance of \methodname{} with ACA and FEA on FewRel under three memory sizes, 5, 10 and 20.

As shown in Table \ref{tab:memory_size}:
(\textit{i}) \methodname{} outperforms strong baselines under all three different memory sizes.
Surprisingly, the performance of \methodname{} with $\mathcal{M}$=10 even defeats baselines with $\mathcal{M}$=20, showing the superior performance of our method.
(\textit{ii})
\methodname{} consistently increases performance than FEA, the two-stage training backbone of our method.
Specifically, \methodname{} outperforms FEA by $1.5$, $1.3$ and $1.4$ in accuracy when the memory size is $5$, $10$, $20$, which indicates that our methods can reduce the dependence on the memory size.

\begin{table}[t]
\centering
\small
\begin{tabular}{lccc}
\toprule 
\textbf{Models} & $\mathcal{M}$=5 & $\mathcal{M}$=10 & $\mathcal{M}$=20 \\
\midrule
FEA & 81.8 & 84.3 & 85.0 \\
ACA  & 82.7 & 84.7 & 85.5\\
\midrule
\methodname{} & 83.3 & 85.6 & 86.4 \\
\bottomrule
\end{tabular}
\caption{Model performance on FewRel under different memory sizes. Our method has lighter dependence on memory size.
}
\label{tab:memory_size}
\end{table}



\section{Conclusion}
\label{sec:bibtex}

In this paper, we focus on continual learning for text classification in the class-incremental setting.
We formally analyze the representation bias problem in continual learning from an information bottleneck perspective and find that learning more class-relevant features could alleviate the bias.
Based on our analysis, we propose \methodname{}, which exploits both contrastive and generative representation learning objectives to capture more class-relevant features, and uses adversarial replay to better recover old knowledge.
Extensive experiments on three tasks show that \methodname{} learns effective representations and significantly outperforms the latest baselines.


\section*{Limitations}
Our paper has several limitations:
(\textit{i}) Our proposed \methodname{} utilizes contrastive and generative objectives to learn class-relevant representations, which introduces extra computational overhead and is less efficient than other replay-based CL methods;
(\textit{ii}) We only focus on catastrophic forgetting problem in continual text classification. How to encourage knowledge transfer in CL is not explored in this paper.

\section*{Ethics Statement}
Our work complies with the ACL Ethics Policy.
As text classification is a standard task in NLP and all datasets we used are public, we do not see any critical ethical considerations.

\bibliography{anthology,custom}
\bibliographystyle{acl_natbib}

\clearpage

\appendix




\section{Details about SupInfoNCE}
\label{app:supinfonce}





\subsection{Connection with InfoNCE}

InfoNCE \citep{cpc} is commonly used in multi-view contrastive representation learning as a proxy to approximate mutual information:
\begin{equation}
\resizebox{.95\hsize}{!}{$
\begin{aligned}
&\mathcal{L}_{\mathrm{InfoNCE}}= \\
&- \log \frac{\exp (f(x)\cdot f(x^+) /\tau)}{\exp (f(x)\cdot f(x^+) /\tau) + \sum_{x^-} \exp (f(x)\cdot f(x^-) /\tau)},
\end{aligned}
$}
\end{equation}
where $f$ is the encoder, $x^+$ is a positive view of $x$, $x^-$ is negative instance sampled from the whole input space $q(x^-)$, $x^-\neq x_i^+$, and $\tau$ is the temperature hyperparameter.

In Section \ref{sec:obj}, we propose to maximize the MI between representations of instances from the same class.
Under our scenario, a batch may contain several positive samples with the same label as the anchor sample.
Since vanilla InfoNCE loss only considers the case of only one positive sample, it should be adapted to a multiple-positive version.
\citet{supinfonce} derive a multiple-positive extension of InfoNCE, SupInfoNCE:
\begin{equation}
\resizebox{.95\hsize}{!}{$
\begin{aligned}
&\mathcal{L}_{\mathrm{SupInfoNCE}}= \\
&-\sum_i \log \frac{\exp (f(x)\cdot f(x^+_i) /\tau)}{\exp (f(x)\cdot f(x^+_i) /\tau) + \sum_{x^-} \exp (f(x)\cdot f(x^-) /\tau)},
\end{aligned}
$}
\end{equation}
where $x_i^+$ is a positive instance with the same label as $x$ in current batch.

Although both InfoNCE and SupInfoNCE are lower bounds of mutual information, due to the difference of positive instances, they are actually approximating difference objectives.
For InfoNCE, since $x^+$ is another view of instance $x$, it will pull representations of the same instance together and push apart representations of all other instances and optimizing InfoNCE is maximizing the lower bound of MI $I(X;Z)$ between input and representation \citep{poole}.
In contrast, $x^+_i$ and $x$ in SupInfoNCE are instances from the same class, optimizing $I(Z;Z^+)$ is pulling representations of instances of the same class together, which instructs the model capture more class-relevant features.

\subsection{Comparison with SupCon}
Previous work \citep{zhao-etal-2022-consistent} uses SupCon \citep{supcon}, a popular supervised contrastive loss, to conduct contrastive learning:
\begin{equation}
\resizebox{.95\hsize}{!}{$
\begin{aligned}
&\mathcal{L}_{\mathrm{SupCon}}= \\
&-\sum_i \log \frac{\exp (f(x)\cdot f(x^+_i) /\tau)}{\sum_i\exp (f(x)\cdot f(x^+_i) /\tau) + \sum_{x^-} \exp (f(x)\cdot f(x^-) /\tau)},
\end{aligned}
$}
\end{equation}
However, \citet{supinfonce} find that SupCon contains a non-contrastive constraint on the positive samples, which may harm the representation learning performance.
Moreover, the connection between SupCon and mutual information has not been well studied.
Therefore, we use SupInfoNCE instead.

\section{Details about FreeLB}
\label{app:adv}
In memory replay stage, to alleviate the overfitting on memorized instances, we introduce FreeLB \citep{freelb} adversarial loss:
\begin{equation}
\label{eq:app_freelb}
\resizebox{.9\hsize}{!}{$
\begin{aligned}
&\mathcal{L}_{\mathrm{adv}}= \\
&\min\limits_{\theta} \mathbb{E}_{(x,y)\sim \mathcal{M}} \left[ \frac{1}{K}\sum_{t=0}^{K-1} \max\limits_{\Vert\delta_t\Vert \le \epsilon} \mathcal{L}\left( F(x+\delta_t),y \right) \right],
\end{aligned}
$}
\end{equation}
where $F$ is the text classification model and $(x,y)$ is a batch of data from the memory bank $\mathcal{M}$, $\delta$ is the perturbation constrained within the $\epsilon$-ball, $K$ is step size hyperparameter. 
The inner maximization problem in \eqref{eq:app_freelb} is to find the worst-case adversarial examples to maximize the training loss, while the outer minimization problem in \eqref{eq:app_freelb} aims at optimizing the model to minimize the loss of adversarial examples.
The inner maximization problem is solved iteratively in FreeLB:
\begin{gather}
\nabla(\delta_{t-1})=\nabla_\delta \mathcal{L}\left(F(x+\delta_{t-1}),y \right), \\
\delta_{t} = \prod_{\Vert\delta\Vert\le \epsilon}\left( \delta_{t-1} + \alpha\cdot \frac{\nabla(\delta_{t-1})}{\Vert \nabla(\delta_{t-1})\Vert} \right),
\end{gather}
where $\delta_{t}$ is the perturbation in $t$-th step and $\prod_{\Vert\delta\Vert\le \epsilon}(\cdot)$ projects the perturbation onto the $\epsilon$-ball, $\alpha$ is step size.

Intuitively, FreeLB performs multiple adversarial attack iterations to craft adversarial examples,
and simultaneously accumulates the free parameter gradients $\nabla_\theta \mathcal{L}$ in each iteration.
After that, the model parameter $\theta$ is updated all at once with the accumulated gradients,
which is equivalent to replacing the original batch with a $K$-times larger adversarial augmented batch.

\section{Dataset Details}
\label{app:dataset}

\paragraph{FewRel}\citep{han-etal-2018-fewrel} It is a large scale relation extraction dataset containing 80 relations. 
FewRel is a balanced dataset and each relation has 700 instances.
Following \citet{zhao-etal-2022-consistent,aca}, we merge the original train and valid set of FewRel and for each relation we sample 420 instances for training and 140 instances for test.
FewRel is licensed under MIT License.

\paragraph{TACRED}\citep{zhang-etal-2017-position} It is a crowdsourcing relation extraction dataset containing 42 relations (including \textit{no\_relation}) and 106264 instances.
Following \citet{zhao-etal-2022-consistent,aca}, we remove \textit{no\_relation} and in our experiments.
Since TACRED is a imbalanced dataset, for each relation the number of training instances is limited to 320 and the number of test instances is limited to 40.
TACRED is licensed under LDC User Agreement for Non-Members.

\paragraph{MAVEN}\citep{wang-etal-2020-maven} It is a large scale event detection dataset with 168 event types.
Since MAVEN has a severe long-tail distribution, we use the data of the top 120 frequent classes.
The original test set of MAVEN is not publicly available, and we use the original development set as our test set.
MAVEN is licensed under Apache License 2.0.

\paragraph{HWU64}\citep{hwu64} It is an intent classification dataset with 64 intent classes.
Following \citet{msr}, we use the data of the top 50 frequent classes and the total number of instances are 24137.
HWU64 is licensed under CC-BY-4.0 License.

\section{Forgetting Rate}
\label{app:forgetting_rate}

Followed \citet{rwalk}, forgetting for a particular task is defined as the difference between the maximum knowledge gained about the task throughout the learning process in the past and the knowledge the model currently has about it.
More concretely, for a classification problem, after training on task $j$, we denote $a_{j,i}$ as the accuracy evaluated on the test set of task $i\le j$:
Then the forgetting rate at $k$-th task is computed as:
\begin{gather}
f_i^k=\max_{l\in\{1,...,k-1\}}a_{l,i}-a_{k,i} \\
\mathrm{FR}_k=\frac{1}{k-1}\sum_{i=1}^{k-1} f_i^k.
\end{gather}
We denote FR as the forgetting rate after finishing all tasks.
Lower FR implies less forgetting on previous tasks.

\begin{table*}
    \centering
    \begin{tabular}{lccccccccc}
    \toprule
    \textbf{Dataset} & $\lambda_1$ & $\lambda_2$ & $Q$ & $\tau$ & $\eta$ & $p_{\mathrm{xmlm}}$ & $K$ & $\alpha$ & $\epsilon$ \\
    \midrule
    FewRel & 0.05 & 0.2 & 512 & 0.05 & 0.99 & 0.5 & 2 & 0.1 & 0.2 \\
    TACRED & 0.05 & 0.05 & 1024 & 0.1 & 0.99 & 0.2 & 2 & 0.1 & 0.2 \\
    MAVEN & 0.05 & 0.05 & 512 & 0.05 & 0.99 & 0.1 & 2 & 0.1 & 0.2 \\
    HWU64 & 0.05 & 0.1 & 512 & 0.05 & 0.99 & 0.4 & 2 & 0.1 & 0.2 \\
    \bottomrule
    \end{tabular}
    \caption{Details of hyperparameters of \methodname{} in each dataset: contrastive loss weighting coefficient $\lambda_1$, XMLM loss coefficient $\lambda_2$, contrastive queue size $Q$, contrastive loss temperature $\tau$, exponential moving average decay rate $\eta$, masked proportion $p_{\mathrm{xmlm}}$, adversarial steps $K$, adversarial step size $\alpha$, adversarial maximum pertubation $\epsilon$. }
    \label{tab:hyp}
\end{table*}

\section{Baselines}
\label{app:baselines}
\textbf{IDBR} \citep{huang-etal-2021-continual} proposes an information disentanglement method to learn representations that can well generalize to future tasks.
\textbf{KCN} \citep{cao-etal-2020-incremental} utilizes prototype retrospection and hierarchical distillation to consolidate knowledge.
\textbf{KDRK} \citep{yu-etal-2021-lifelong} encourages knowledge transfer between old and new classes.
\textbf{EMAR} \citep{han-etal-2020-continual} proposes a memory activation and reconsolidation mechanism to retain the learned knowledge.
\textbf{RP-CRE} \citep{cui-etal-2021-refining} proposes a memory network to retain the learned representations with class prototypes.
\textbf{CRL} \citep{zhao-etal-2022-consistent} adopts contrastive learning replay and knowledge distillation to retain the learned knowledge.
\textbf{CRECL} \citep{hu-etal-2022-improving} uses a prototypical contrastive network to defy forgetting.
\textbf{FEA} \citep{wang2022less} demonstrates the effectiveness of two-stage training framework, which can be seen as the backbone of \methodname{}.
\textbf{ACA} \citep{aca} designs two adversarial class augmentation mechanism to learn robust representations.

\section{Implementation Details}
\label{app:details}

We implement \methodname{} with PyTorch \citep{pytorch} and HuggingFace Transformers \citep{wolf-etal-2020-transformers}.
Following previous work \cite{cui-etal-2021-refining, aca, wang2022less}, we use BERT$_{\mathrm{base}}$ \citep{devlin-etal-2019-bert} as encoder.
PyTorch is licensed under the modified BSD license.
HuggingFace Transformers and BERT$_{\mathrm{base}}$ are licensed under the Apache License 2.0.
Our use of existing artifacts is consistent with their intended use.

The text classification model can be seen as two parts: the BERT encoder $f_\theta$, and the MLP classifier $g_\phi$.
Specifically, for the input $x$ in relation extraction, we use $[E_{11}],\ [E_{12}],\ [E_{21}]$ and $[E_{22}]$ to denote the start and end position of head and tail entity respectively, and the representation $z$ is the concatenation of the last hidden states of $[E_{11}]$ and $[E_{21}]$.
For the input $x$ in event detection, the representation $z$ is the average pooling of the last hidden states of the trigger words.
For the input $x$ in intent detection, the representation $z$ is the average pooling of the last hidden states of the whole sentence.

The learning rate of \methodname{} is set to 1e-5 for the BERT encoder and 1e-3 for the other layers.
The training epochs for both the initial learning stage and the memory replay are 10.
For FewRel, MAVEN and HWU64, the batch size is 32. 
For TACRED, the batch size is 16 because of the small amount of training data.
The budget of memory bank for each class is 10 for all methods.
We tune the hyperparameters of \methodname{} on the first three tasks for each dataset.
Details of hyperparameter setting of \methodname{} are shown in Table \ref{tab:hyp}.
For all experiments, we use NVIDIA A40 and RTX 3090 GPUs and report the average result of 5 different task sequences.

\section{Full Experiment Results}
\label{app:main_exp}
The full experiment results on 10 continual learning tasks are shown in Table \ref{tab:full}.
As shown, \methodname{} consistently outperforms baselines in nearly all task stages on four datasets.

\begin{table*}[t]
    \centering
    \scalebox{0.85}{
    \begin{tabular}{lccccccccccc}
    
     \toprule
     \multicolumn{12}{c}{\textbf{FewRel}} \\
     \midrule
      \textbf{Models} & \textbf{T1} & \textbf{T2} & \textbf{T3} & \textbf{T4} & \textbf{T5} & \textbf{T6} & \textbf{T7} & \textbf{T8} & \textbf{T9} & \textbf{T10} & \textbf{FR}  \\
    \midrule
    IDBR \citep{huang-etal-2021-continual} & 97.9 &  91.9 & 86.8 & 83.6 & 80.6 & 77.7 & 75.6 & 73.7 & 71.7 & 68.9 & 30.4 \\
     KCN \citep{cao-etal-2020-incremental} & 98.3 & 93.9 & 90.5 & 87.9 & 86.4 & 84.1 & 81.9 & 80.3 & 78.8 & 76.0 & 23.2 \\
    KDRK \citep{yu-etal-2021-lifelong} & 98.3 & 94.1 & 91.0 & 88.3 & 86.9 & 85.3 & 82.9 & 81.6 & 80.2 & 78.0 & 18.4 \\
    EMAR \citep{han-etal-2020-continual} & 98.1 & 94.3 & 92.3 & 90.5 & 89.7 & 88.5 & 87.2 & 86.1 & 84.8 & 83.6 & 12.1 \\
    RP-CRE \citep{cui-etal-2021-refining} & 97.8 & 94.7 & 92.1 & 90.3 & 89.4 & 88.0 & 87.1 & 85.8 & 84.4 & 82.8 & 10.3 \\
    CRL \citep{zhao-etal-2022-consistent} & 98.2 & 94.6 & 92.5 & 90.5 & 89.4 & 87.9 & 86.9 & 85.6 & 84.5 & 83.1 & 11.5 \\
    CRECL \citep{hu-etal-2022-improving} & 97.8 & 94.9 & 92.7 & 90.9 & 89.4 & 87.5 & 85.7 & 84.6 & 83.6 & 82.7 & 11.6 \\
    FEA \citep{wang2022less} & {98.3} & {94.8} & {93.1} & {91.7} & {90.8} & {89.1} & {87.9} & {86.8} & {85.8} & {84.3} & 8.9  \\
    ACA \citep{aca} & 98.3 & 95.0 & 92.6 & 91.3 & 90.4 & 89.2 & 87.6 & 87.0 & 86.3 & 84.7 & 11.0 \\
    \midrule
    \methodname{}  & 98.2 &  95.3 & 93.6 & 92.4 & 91.6 & 90.3 & 88.8 & 88.1 & 86.9 & 85.6 & 8.7 \\
     \bottomrule

     \toprule
     \multicolumn{12}{c}{\textbf{TACRED}} \\
     \midrule
      \textbf{Models} & \textbf{T1} & \textbf{T2} & \textbf{T3} & \textbf{T4} & \textbf{T5} & \textbf{T6} & \textbf{T7} & \textbf{T8} & \textbf{T9} & \textbf{T10} & \textbf{FR}  \\
    \midrule
     IDBR \citep{huang-etal-2021-continual} & 97.9 & 91.1 & 83.1 & 76.5 & 74.2 & 70.5 & 66.6 & 64.2 & 63.8 & 60.1 & 35.3 \\
     KCN \citep{cao-etal-2020-incremental} & 98.9 & 93.1 & 87.3 & 80.2 & 79.4 & 77.2 & 73.8 & 72.1 & 72.2 & 70.6 & 22.3 \\
    KDRK \citep{yu-etal-2021-lifelong} & 98.9 & 93.0 & 89.1 & 80.7 & 79.0 & 77.0 & 74.6 & 72.9 & 72.1 & 70.8 & 22.8 \\
    EMAR \citep{han-etal-2020-continual} & 98.3 & 92.0 & 87.4 & 84.1 & 82.1 & 80.6 & 78.3 & 76.6 & 76.8 & 76.1 & 20.0 \\
    RP-CRE \citep{cui-etal-2021-refining} & 97.5 & 92.2 & 89.1 & 84.2 & 81.7 & 81.0 & 78.1 & 76.1 & 75.0 & 75.3 & 17.5 \\
    CRL \citep{zhao-etal-2022-consistent} & 97.7 & 93.2 & 89.8 & 84.7 & 84.1 & 81.3 & 80.2 & 79.1 & 79.0 & 78.0 & 18.0 \\
    CRECL \citep{hu-etal-2022-improving} & 96.6 & 93.1 & 89.7 & 87.8 & 85.6 & 84.3 & 83.6 & 81.4 & 79.3 & 78.5 & 16.3 \\
    FEA \citep{wang2022less}& 97.6 & 92.6 & 89.5 & {86.4} & {84.8} & {82.8} & {81.0} & 78.5 & 78.5 & 77.7 & 13.3 \\
    ACA \citep{aca} & 98.0 & 92.1 & 90.6 & 85.5 & 84.4 & 82.2 & 80.0 & 78.6 & 78.8 & 78.1 & 13.8 \\
    \midrule
    \methodname{} & 98.0 & 93.3 & 91.7 & 87.8 & 85.4 & 83.6 & 81.5 & 79.9 & 79.8 & 78.6 & 12.0  \\
     \bottomrule

     \toprule
     \multicolumn{12}{c}{\textbf{MAVEN}} \\
     \midrule
      \textbf{Models} & \textbf{T1} & \textbf{T2} & \textbf{T3} & \textbf{T4} & \textbf{T5} & \textbf{T6} & \textbf{T7} & \textbf{T8} & \textbf{T9} & \textbf{T10} & \textbf{FR}  \\
    \midrule
     IDBR \citep{huang-etal-2021-continual} & 96.5 & 85.3 & 79.4 & 76.3 & 74.2 & 69.8 & 67.5 & 64.4 & 60.2 & 57.3 & 34.2 \\
     KCN \citep{cao-etal-2020-incremental} & 97.2 & 87.7 & 83.2 & 80.3 & 77.9 & 75.1 & 71.9 & 68.4 & 67.7 & 64.4 & 29.0 \\
    KDRK \citep{yu-etal-2021-lifelong} & 97.2 & 88.6 & 84.3 & 81.6 & 78.1 & 75.8 & 72.5 & 69.6 & 68.9 & 65.4 & 28.3 \\
    EMAR \citep{han-etal-2020-continual} & 97.2 & 91.4 & 88.3 & 86.1 & 83.6 & 81.2 & 79.0 & 76.8 & 75.7 & 73.2 & 14.6 \\
    RP-CRE \citep{cui-etal-2021-refining} & 96.6 & 92.1 & 88.6 & 86.7 & 83.9 & 82.0 & 79.4 & 77.2 & 77.0 & 74.8 & 11.4 \\
    CRL \citep{zhao-etal-2022-consistent} & 96.0 & 90.7 & 87.1 & 84.8 & 82.9 & 80.7 & 78.7 & 76.8 & 75.9 & 73.7 & 11.2 \\
    CRECL \citep{hu-etal-2022-improving} & 96.9 & 91.4 & 86.9 & 84.8 & 82.4 & 80.4 & 77.5 & 75.9 & 75.1 & 73.5 & 13.8 \\
    FEA \citep{wang2022less} & 97.2 & 92.0 & 88.6 & 86.2 & 84.4 & 82.1 &  79.7 & 78.0 & 77.0 & 75.0 & 12.8   \\
    \midrule
    \methodname{} &97.1&92.2&88.9&86.7&84.8&82.5&80.4&78.4&77.7&75.9 & 10.7  \\
     \bottomrule

     \toprule
     \multicolumn{12}{c}{\textbf{HWU64}} \\
     \midrule
      \textbf{Models} & \textbf{T1} & \textbf{T2} & \textbf{T3} & \textbf{T4} & \textbf{T5} & \textbf{T6} & \textbf{T7} & \textbf{T8} & \textbf{T9} & \textbf{T10} & \textbf{FR}  \\
    \midrule
     IDBR \citep{huang-etal-2021-continual} & 96.3 & 93.2 & 88.1 &  86.5 & 84.6 & 82.5 & 82.1 & 80.2 & 78.0 & 76.2 & 19.0 \\
     KCN \citep{cao-etal-2020-incremental} & 98.6 & 94.0 & 90.7 & 90.4 & 87.0 & 84.9 & 84.4 & 83.7 & 82.7 & 81.9 & 13.5 \\
    KDRK \citep{yu-etal-2021-lifelong} & 98.6 & 94.5 & 91.2 & 90.4 & 87.3 & 86.0 & 85.8 & 85.1 & 82.5 & 81.4 & 14.0 \\
    EMAR \citep{han-etal-2020-continual} & 98.4 & 94.4 & 91.4 & 89.5 & 88.2 & 86.3 & 86.2 & 85.5 & 83.9 & 83.1 & 9.3 \\
    RP-CRE \citep{cui-etal-2021-refining} & 97.6 & 93.7 & 90.1 & 88.6 & 86.5 & 86.3 & 85.1 & 84.5 & 83.8 & 82.7 & 10.9 \\
    CRL \citep{zhao-etal-2022-consistent} & 98.2 & 92.8 & 88.8 & 86.5 & 84.1 & 82.4 & 82.8 & 83.1 & 81.3 & 81.5 & 9.9 \\
    CRECL \citep{hu-etal-2022-improving} & 97.3 & 93.0 & 87.5 & 86.1 & 84.1 & 83.0 & 83.1 & 83.1 & 81.9 & 81.1 & 9.8 \\
    FEA \citep{wang2022less}  & 96.5	& 94.4 & 90.0	& 89.7	& 88.3	& 87.2	& 86.0	& 85.4  &84.0 &	83.3 & 8.8 \\
    \midrule
    \methodname{} &  97.7 & 93.7 & 91.5 & 89.9 & 88.9 & 87.4 & 87.5 & 87.0 & 84.9 & 84.8 & 8.0 \\
     \bottomrule
     
    \end{tabular}}
    \caption{
    Accuracy (Acc, \%) on all observed classes at at the stage of learning current task and forgetting rate (FR) after learning all tasks.
    Note some baselines are originally proposed to tackle one specific task.
    We adapt these baselines on other tasks and report the corresponding results.
    \methodname{} consistently outperforms baselines in nearly all task stages on four datasets.
    }
    \label{tab:full}
\end{table*}

\end{document}